\documentclass[letterpaper,10pt,journal,twoside]{IEEEtran}

\IEEEoverridecommandlockouts  

\usepackage{microtype}
\usepackage{amsmath, amssymb}
\usepackage{graphicx}
\usepackage{stfloats}
\usepackage[ruled,vlined,noend]{algorithm2e}
\usepackage{booktabs, multirow}
\usepackage{balance}
\usepackage[colorlinks]{hyperref}

\usepackage{pifont}
\newcommand{\cmark}{\ding{51}}
\newcommand{\xmark}{\ding{55}}

\title{mBEST: Realtime Deformable Linear Object Detection Through Minimal Bending Energy Skeleton Pixel Traversals}

\author{Andrew Choi$^{1}$, Dezhong Tong$^{2}$, Brian Park$^{1}$, \\ Demetri Terzopoulos$^{1}$, Jungseock Joo$^{3}$, and Mohammad Khalid Jawed$^{\dagger, 2}$
\thanks{Manuscript received: February, 18, 2023; Revised May, 22, 2023; Accepted June, 12, 2023. This paper was recommended for publication by Editor Cesar Cadena upon evaluation of the Associate Editor and Reviewers' comments.
This work was supported by the National Science Foundation under award numbers OAC-2209782, CMMI-2101751, CAREER-2047663, and IIS-1925360.} 
\thanks{The authors are with the University of California, Los Angeles (UCLA), Los Angeles, CA 90095, USA}%
\thanks{$^{1}$Andrew Choi, Brian Park, and Demetri Terzopoulos are with the UCLA Computer Science Department (email: {\tt\footnotesize asjchoi@cs.ucla.edu; briannparkk@ucla.edu; dt@cs.ucla.edu}).}%
\thanks{$^{2}$Dezhong Tong and Mohammad Khalid Jawed are with the UCLA Department of Mechanical \& Aerospace Engineering (email: {\tt \footnotesize tltl960308@g.ucla.edu; khalidjm@seas.ucla.edu}).}%
\thanks{$^{3}$Jungseock Joo is with the UCLA Department of Communication and is currently working at NVIDIA Corporation (email: {\tt \footnotesize jjoo@comm.ucla.edu}).}%
\thanks{$^\dagger$ Corresponding author.}%
\thanks{Digital Object Identifier 10.1109/LRA.2023.3290419}
}

\begin{document}
\maketitle

\begin{abstract}
Robotic manipulation of deformable materials is a challenging task that often requires realtime visual feedback. 
This is especially true for deformable linear objects (DLOs) or ``rods'', whose slender and flexible structures make proper tracking and detection nontrivial. 
To address this challenge, we present \textit{mBEST}, a robust algorithm for the realtime detection of DLOs that is capable of producing an ordered pixel sequence of each DLO\rq{}s centerline along with segmentation masks.
Our algorithm obtains a binary mask of the DLOs and then thins it to produce a skeleton pixel representation.
After refining the skeleton to ensure topological correctness, the pixels are traversed to generate paths along each unique DLO.
At the core of our algorithm, we postulate that intersections can be robustly handled by choosing the combination of paths that minimizes the cumulative bending energy of the DLO(s). 
We show that this simple and intuitive formulation outperforms the state-of-the-art methods for detecting DLOs with large numbers of sporadic crossings ranging from curvatures with high variance to nearly-parallel configurations.
Furthermore, our method achieves a significant performance improvement of approximately 50\% faster runtime and better scaling over the state of the art.
\end{abstract}


\section{Introduction}
\label{sec:introduction}

\IEEEPARstart{A}S robots become increasingly more intelligent and capable, developing robust and effective deformable material manipulation skills has started to attract substantial research attention~\cite{sanchez2018robotic}. 
Among various deformable objects, deformable linear objects (DLOs) --- typically referred to as ``rods'' by the mechanics community --- are a special group, including everyday objects such as cables, ropes, tubes, and threads. 
Due to their distinctive geometric characteristic (width $\approx$ height $\ll$ length), DLOs are widely used in various domestic and industrial applications, including surgical suturing~\cite{schulman2013case}, knot fastening~\cite{mayer2008system, choi2021implicit}, cable manipulation~\cite{zhu2018dual, yu2022shape}, food manipulation~\cite{iqbal2017prospects}, mechanics analysis~\cite{tong2021automated}, and more. 
Because of their flexibility, DLOs are often prone to complex tangling, which complicates manipulation. 
Additionally, the complicated structures made by DLOs usually have unique topology-induced mechanical properties~\cite{audoly2007elastic, jawed2015untangling, patil2020topological, johanns2021shapes, sano2022exploring} and are, therefore, used to tie knots for sailing, fishing, climbing, and various other engineering applications.
Given all the aforementioned, a robust, efficient, and accurate perception algorithm for DLOs is crucial to both deformable material manipulation and soft robotics.

We present an algorithm for robust, accurate, and fast instance segmentation of DLOs, named \textit{mBEST} (Minimal Bending Energy Skeleton pixel Traversals).
Without any prior knowledge regarding the geometries, colors, and total number of DLOs, \textit{mBEST} takes a raw RGB image as input and outputs a series of ordered pixels defining the centerline of each individual DLO in the image, thus allowing for the configurations of different DLOs to be easily incorporated into motion planning and manipulation schemes.

\begin{figure*}
	\includegraphics[width=\textwidth]{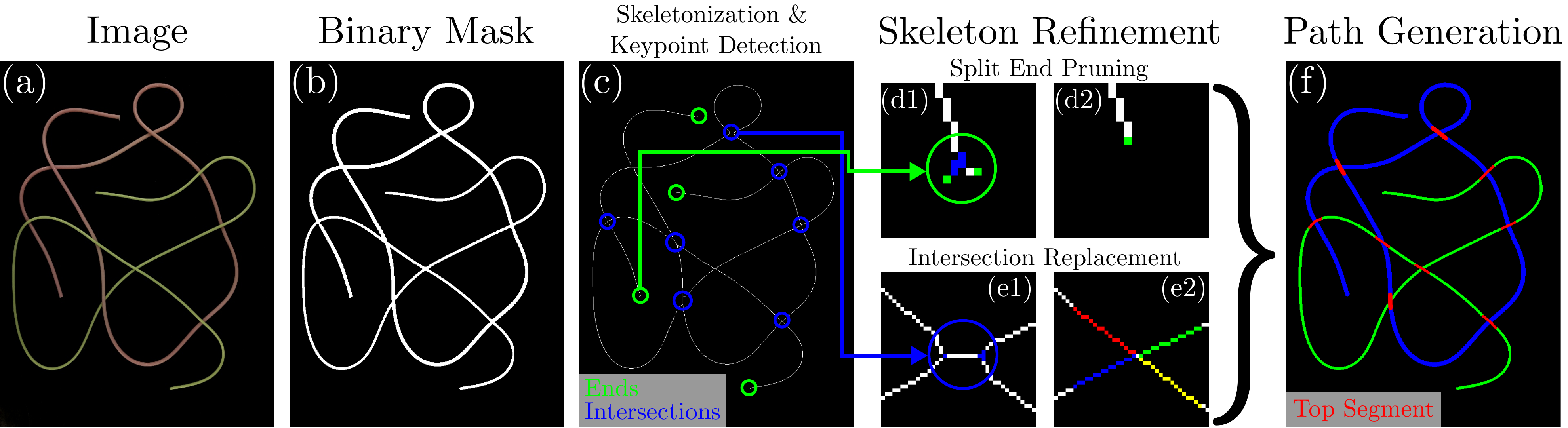}
	\caption{Overview of the \textit{mBEST} processing pipeline. An input image (a) is converted to a binary mask (b) using a segmentation method. The binary mask is then converted to a skeleton pixel representation (c), where the connectivity and centerlines of the DLOs are preserved as a single-pixel wide structure and keypoints, such as intersections and ends, are detected. This is followed by a series of refinement steps to maintain the topological correctness of the skeleton: split ends (d1) are pruned (d2) and pixels representing a single topological intersection (e1) are clustered, matched, and replaced with a more intuitive intersection (e2). Finally, the DLOs are delineated (f) by traversing skeleton pixels and choosing minimal cumulative bending energy paths.} 
\label{fig:pipeline}
\end{figure*}

To achieve instance segmentation of DLOs in images, we implement the following sequence of processing steps: 
Like previous work \cite{caporali2022fastdlo}, we first perform semantic segmentation to produce a binary mask of the DLOs against the background using either simple color filtering methods or a Deep Convolutional Neural Network (DCNN).
After a binary mask is obtained, we apply a thinning algorithm to the mask to produce a single-pixel-wide skeleton representation of the DLOs, which preserves the connectivity and centerlines of the binary mask.
Thus, key points such as ends and intersections are easily detected. 
After a series of refinement steps to ensure topological correctness, the skeleton is then traversed, one end at a time, in a manner that minimizes the cumulative bending energy of the DLOs, until another end is encountered.
Each traversal yields a single DLO's centerline pixel coordinates, which optionally can then be used to produce segmentation masks.
Fig.~\ref{fig:pipeline} overviews the \textit{mBEST} processing pipeline.

Overall, our main contributions in this article are that we
\begin{enumerate}
    \item develop a robust pipeline for obtaining ordered centerline coordinates and segmentation masks of DLOs from semantic binary masks;
    \item demonstrate that the relatively simple and physically meaningful optimization objective of minimizing cumulative bending energy outperforms several state of the art (SOTA) algorithms;
    \item showcase the effectiveness of our topology-correcting skeleton refinement steps by outperforming the SOTA algorithms with a hybrid \textit{mBEST} formulation that uses the intersection handling scheme of SOTA algorithms;
    \item achieve faster, real-time performance compared to the SOTA algorithms.
\end{enumerate}
Moreover, we have released all our source code, datasets (with ground truth), and a supplementary video.\footnote{\label{github} See \url{https://github.com/StructuresComp/mBEST}.}

The remainder of the article is organized as follows:
We present a review of related work in Sec.~\ref{sec:related_work}.
The algorithmic formulation of \textit{mBEST} is then detailed in Sec.~\ref{sec:methodology}.
In Sec.~\ref{sec:exp}, we report our experimental results comparing \textit{mBEST} with the SOTA approaches.
Finally, we make concluding remarks and discuss potential future research directions in Sec.~\ref{sec:conclusion}.

\section{Related Work}
\label{sec:related_work}

Although research into manipulation skills for DLOs has been prevalent, the perception algorithms used in support of these efforts remain underdeveloped. 
For example, in the work of Tong et al.~\cite{tong2021automated}, attached markers are required to determine the configuration of the manipulated DLO.
Zhu et al.~\cite{zhu2018dual} carefully adjusted the workspace to increase the contrast between the manipulated DLOs (cables) and their background. 
Although these prior efforts successfully completed their target manipulation tasks, the simplistic perception algorithms restrict real world applicability.

Consequently, DLO detection algorithms featuring various methodologies have been proposed. 
Keipour et al.~\cite{keipour2022deformable} evaluated both curvatures and distances to fit a continuous DLO. 
Using data-driven methods, Yan et al.~\cite{yan2020self} trained a neural network to reconstruct the topology of a DLO based on a coarse-to-fine nodal representation. 
Though these methods achieve good results for some datasets, they work under the strict assumption that only one DLO exists within the scene, which dramatically restricts their applicability.

One of the first perception algorithms capable of detecting multiple DLOs, \textit{Ariadne}~\cite{gregorio2018superpixels}, segments images into superpixels and traverses the superpixels belonging to DLOs in order to produce paths. 
The ambiguity of intersections is handled using a multi-faceted cost function that takes into consideration color, distance, and curvature.
Despite its satisfactory performance, this early approach suffers from a large number of hyperparameters, an overreliance on DLOs being a uniform color, and the tedious requirement that the user manually select the ends of DLOs.
Furthermore, the processing speed of \textit{Ariadne} is on the order of seconds, precluding realtime operation.

In recent years, data-driven computer vision methods have attracted increasing attention and researchers have shown that image segmentation problems can be tackled efficiently and accurately using Deep Convolutional Neural Networks (DCNNs), particularly instance segmentation~\cite{bolya2019yolact, bolya2020yolact++, chen2020blendmask, tian2020conditional}. 
Furthermore, techniques have been introduced to help synthetically generate large quantities of photorealistic data in order to adequately train such models~\cite{denninger2019blenderproc, qiu2016unrealcv, caporali2023weaksuper}. 
Using DCNNs, Zanella et al.~\cite{zanella2021auto} created segmentations of DLOs such as wires; however, the segmentations did not distinguish between each DLO. 

Improving upon \textit{Ariadne}, \textit{Ariadne+}~\cite{caporali2022ariadne+} also utilizes a DCNN model to extract an initial binary mask of the DLOs.
This allows the algorithm to then apply superpixel segmentation purely on the binary mask itself, significantly reducing the computation time.
Paths are then generated in a similar fashion to the original \textit{Ariadne} algorithm by traversing superpixels while intersections are handled using a neural network to predict the most probable paths.
Despite these improvements, \textit{Ariadne+} is sub-realtime; i.e., less than 3 FPS.

Another algorithm, \textit{FASTDLO}~\cite{caporali2022fastdlo} improves upon the speed of \textit{Ariadne+} by forgoing superpixel segmentation altogether. 
Instead, it uses a skeleton pixel representation of the DLO binary mask for path traversals.
Intersections are then also handled by a neural network.
By replacing superpixel segmentation with skeletonization, \textit{FASTDLO} is able to achieve a realtime performance of 20 FPS for images of size $640 \times 360$ pixels.

More recently, \textit{RT-DLO}~\cite{caporali2023rtdlo} detects DLOs by representing them as sparse graphs where nodes are sampled from DLO centerlines and edges are selected based on topological reasoning. This results in increased runtime efficiency and accuracy compared to \textit{Ariadne+} and \textit{FASTDLO}, but requires sampling along the centerlines of the DLO to remain computationally competitive, often resulting in noisy segmentations.
Furthermore, several hyperparameters must be set.

\textit{Ariadne+}, \textit{FASTDLO}, and \textit{RT-DLO} are considered state-of-the-art DLO perception algorithms, but they have been evaluated only on scenes containing DLOs with relatively smooth curvatures and minimal self-loops.
Our experiments will show that these algorithms struggle to resolve nontrivial configurations (e.g., DLOs with highly variable curvatures resulting in many crossings and tangles and/or nearly-parallel intersections). We argue that a physically principled approach can outperform both sparse graphs and black box neural network approaches when dealing with intersections. Our \textit{mBEST} algorithm robustly solves complex scenes using the simple notion that the most probable path is the one that minimizes cumulative bending energy.
Not only does \textit{mBEST} outperform in accuracy, it also achieves realtime performance with a $50\%$ improvement over the next best algorithm and it has no hyperparameters to set.
Table~\ref{tab:comparison_between_methods} summarizes the key algorithmic differences between \textit{mBEST} and competing algorithms.

\begin{table}
\renewcommand{\arraystretch}{1.2}
\centering
\caption{Comparison of Algorithms}
\begin{tabular}{lccc}
\toprule
\textbf{Algorithm} & \begin{tabular}{@{}c@{}}Intersection \\[-2pt] Rule\end{tabular} &  \begin{tabular}{@{}c@{}}DLO \\[-2pt] Representation\end{tabular} & Realtime \\
\midrule
\textit{Ariadne} \cite{gregorio2018superpixels}  & \begin{tabular}{@{}c@{}}color, distance, \\[-4pt] curvature\end{tabular} & superpixels & \xmark \\
\textit{Ariadne+} \cite{caporali2022ariadne+} & DNN prediction & superpixels & ~~\cmark$^-$ \\
\textit{FASTDLO} \cite{caporali2022fastdlo}  & DNN prediction & skeleton pixels & \cmark \\
\textit{RT-DLO} \cite{caporali2023rtdlo}  & cosine similarity & sparse graph & \cmark \\
\textit{mBEST}  & curvature & skeleton pixels  & \cmark \\
\bottomrule
\end{tabular}
\label{tab:comparison_between_methods}
\end{table}

\section{Methodology}
\label{sec:methodology}

The \textit{mBEST} algorithm consists of the following steps:
\begin{enumerate}
    \item DLO Segmentation
    \item Skeletonization
    \item Keypoint Detection
    \item Split End Pruning
    \item Intersection Clustering, Matching, and Replacement
    \item Minimal Bending Energy Path Generation   
    \item Crossing Order Determination
\end{enumerate}
The following sections describe each step in detail.

\subsection{DLO Segmentation}

The first step in detecting the DLOs is to obtain a binary mask $\mathbf M_\textup{dlo}$ of the image that distinguishes all DLO-related pixels from the background. The initial image segmentation method is not a key contribution of \textit{mBEST}. Rather,  
it is a modular component of our pipeline, allowing for different methods to be plugged in depending on the use case.
As stated previously, we employ two semantic segmentation methods: a DCNN segmentation model and color filtering. 
In particular, we use \textit{FASTDLO}'s pretrained DCNN model~\cite{caporali2022fastdlo} in our experiments.

\subsection{Skeletonization}

As shown in Fig.~\ref{fig:pipeline}(b)--(c), the next step of our algorithm is to convert $\mathbf M_{\textup{dlo}}$ to a skeleton mask $\mathbf M_{\textup{sk}}$, which is useful as both the connectivity and general topology of the DLOs are maintained.
Furthermore, as segments are only 1 pixel wide, traversals along segments are not susceptible to path ambiguity.
To achieve skeletonization, we use an efficient thinning algorithm designed specifically for 2D images, and refer the reader to \cite{zhang1984skeleton} for the details.

\subsection{Keypoint Detection}
\label{subsec:keypoint_detection}

\begin{figure}
	\includegraphics[width=0.48\textwidth]{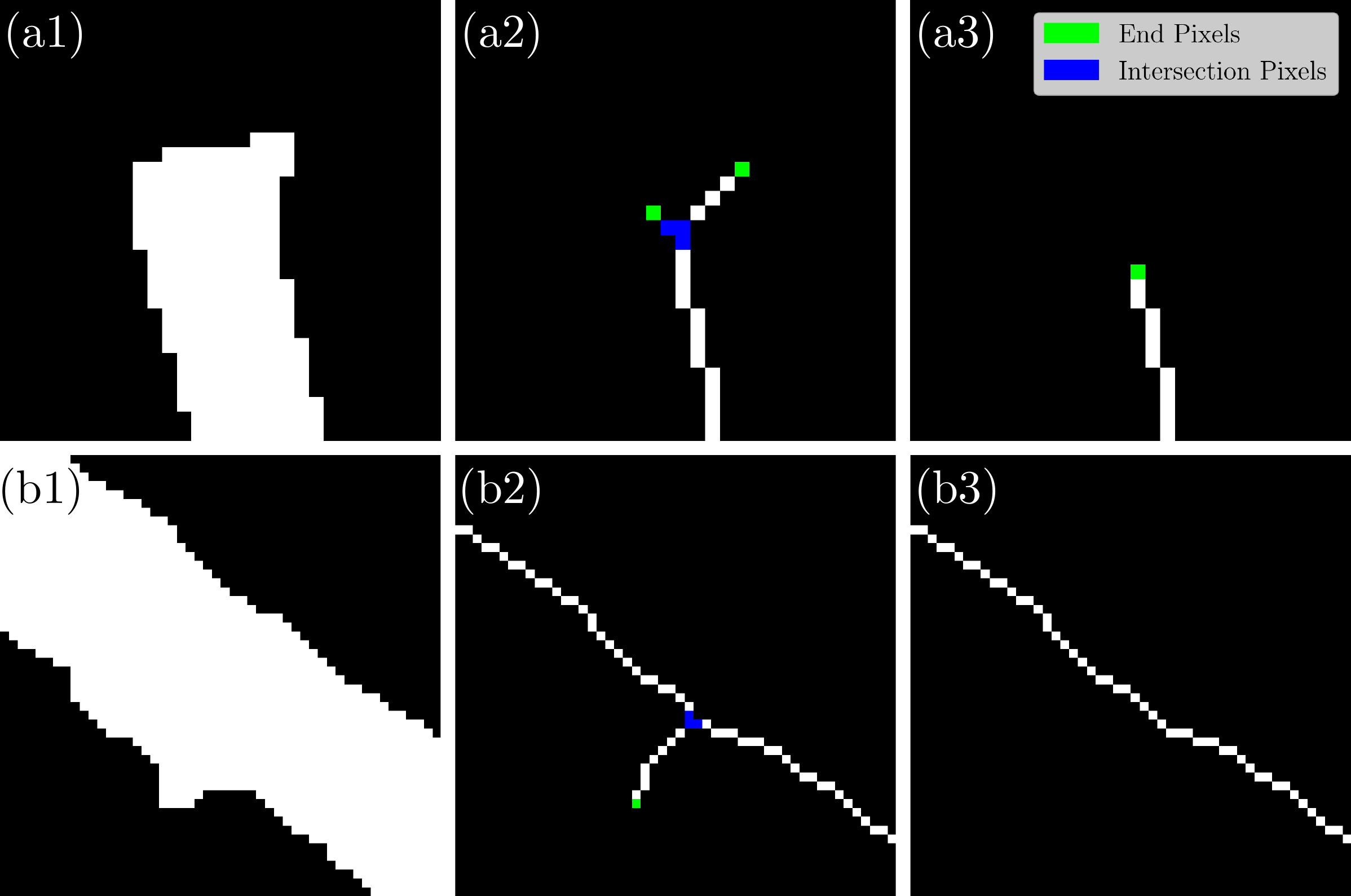}
	\caption{Examples of split ends that may occur during the skeletonization process. Row~(a) shows split ends that may occur at an actual topological end, while Row~(b) shows a split end along a segment produced by a jagged mask. For both examples, the first column shows the binary mask; the second shows the split end after skeletonization, and the third shows the topologically correct structure after pruning.}
	\label{fig:pruning_split_ends}
\end{figure}

After obtaining the skeleton pixel representation, we can then detect two types of key points: ends and intersections.
Locating ends is crucial since they serve as the start and finish points for skeleton pixel traversals.
Locating intersections is crucial as they represent the only points at which a pixel traversal will have multiple possible routes.
Therefore, care must be taken in choosing the correct path when passing through an intersection.

To detect ends and intersections, a skeleton pixel classification kernel,
\begin{equation*}
\mathbf K = 
\begin{bmatrix}
1 & 1  & 1 \\
1 & 10 & 1 \\
1 & 1  & 1
\end{bmatrix},
\label{eq:sk_kernel}
\end{equation*}
is convolved with the skeleton mask; i.e., $\mathbf M_\textup{sk} \circledast \mathbf K$.
We then identify all end pixels $\mathbf E$ as those with a value of 11 (1 neighbor) and all intersection pixels $\mathbf I$ as those with a value greater than 12 (3 or more neighbors).

\begin{figure*}
	\includegraphics[width=\textwidth]{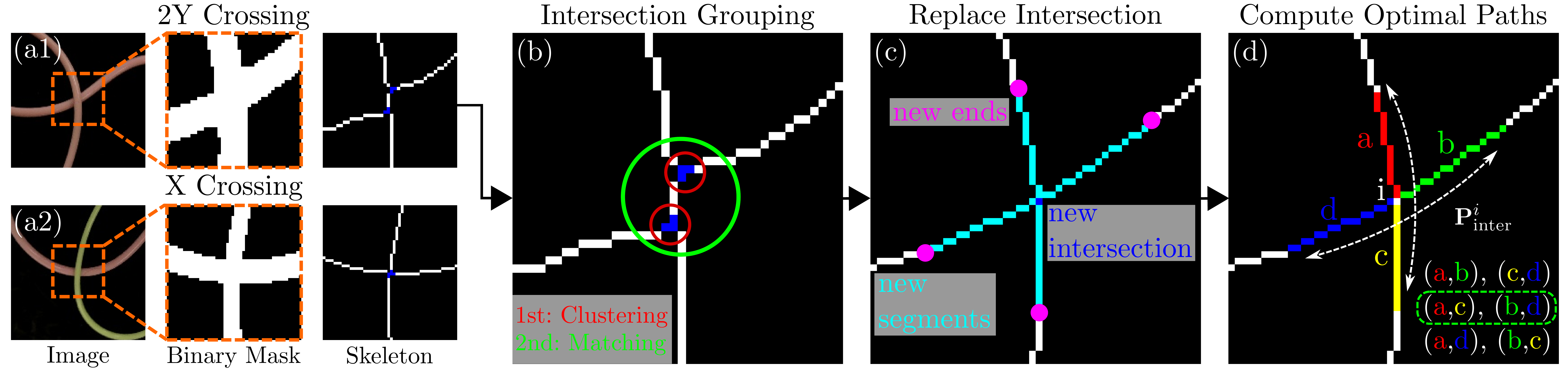}
	\caption{The intersection clustering, matching, replacement, and optimal path generation pipeline. Two sample intersections are shown where skeletonization results in a 2Y-shaped crossing (a1) and an X-shaped crossing (a2). As 2Y-shaped crossings are topologically incorrect, we replace them by replacing the intersection pixels (b) in two stages: the first involves clustering adjacent pixels and the second involves pair matching nearby clusters. Using the centroid location of the matched clusters, we then replace the intersection (c) by creating new ends and having new segments sprout and connect to the centroid. Finally, (d) the new generated ends and segments are used to discover the combination of paths that minimizes the cumulative bending energy of the DLO.}
	\label{fig:intersection_clustering}
\end{figure*}

After obtaining both $\mathbf E$ and $\mathbf I$, additional work must be done to obtain the correct representative sets.
For example, end pixels that are unindicative of a topological end may be produced from a noisy binary mask.
These ``split ends'' will then falsely produce intersection pixels themselves, as shown in Fig.~\ref{fig:pruning_split_ends}.
Additionally, a single topological intersection will result in either two Y-shaped divides or a single X-shaped divide, as shown in Fig.~\ref{fig:intersection_clustering}(a).
Such pixels must be clustered accordingly, with a single point of intersection determined.
In the case of a skeleton possessing two Y-shaped divides in the context of a single intersection, the intersection must also be replaced with an X-shaped divide that more accurately represents the centerlines of the DLOs.

\subsection{Split End Pruning}
\label{subsec:pruning}

When the boundary of the binary mask $\mathbf M_\textup{dlo}$ is jagged, the skeleton mask $\mathbf M_\textup{sk}$ may contain several types of split ends, as shown in Fig.~\ref{fig:pruning_split_ends}.
Such split ends must be identified and pruned as they do not accurately represent the topology of the DLO(s) and will result in incorrect start points as well as cause path ambiguity during pixel traversals.

Note that the length of a split end can be at most the radius of the DLO from which it is sprouting.
Therefore, the length of all split ends should be within a threshold $\delta$ much less than the length of the DLO.
As such, for every end in $\mathbf E$, we traverse along its segment until one of the following three conditions occurs before traversing $\delta$ pixels:
\begin{enumerate}
    \item an intersection is encountered,
    \item an end is encountered,
    \item or neither was encountered.
\end{enumerate}
For Conditions~1 and 2, we remove the segment that was just traversed from $\mathbf M_\textup{sk}$ as well as the corresponding end from $\mathbf E$. 
For Condition~1, we must also remove from $\mathbf I$ all intersection pixels that were produced from the pruned split end.
For any endpoint that satisfies Condition~3, we do nothing. 

To encompass all possible split ends, we can set $\delta$ to be the diameter of the widest DLO in the image. 
Radii of the DLOs can be obtained by computing an L2 distance transform on $\mathbf M_\textup{dlo}$, which results in a matrix $\mathbf D$ containing for each pixel location the closest Euclidean distance to a 0-value pixel.
With this, we can then simply set $\delta =2 \max(\mathbf D)$.
As the distance matrix $\mathbf D$ tells us the radii information for all centerline points, we can reuse it to generate segmentation masks once each DLO's path is ascertained \cite{caporali2022fastdlo}.

\subsection{Intersection Clustering, Matching, and Replacement}
\label{subsec:intersection_clustering}

As mentioned in Sec.~\ref{subsec:keypoint_detection}, a single topological intersection can result in either a 2Y or X-shaped branching as shown in Fig.~\ref{fig:intersection_clustering}(a).
Furthermore, each of these branches may have several intersection pixels; i.e., pixels with 3 or more neighbors.
Our goal then is to group each pixel in $\mathbf I$ to a single branch and then group each branch to its true topological intersection.
With all the intersection pixels properly grouped, we then define a single intersection pixel that represents the true center of a crossing, for all crossings. 

First, to cluster all adjacent intersection pixels, we use Density-Based Spatial Clustering of Applications with Noise (DBSCAN) \cite{ester1996dbscan}, an algorithm that clusters data points within a distance threshold of each other.  
In our case, the Euclidean pixel distance and is simply set to 2.
Once all adjacent pixels in $\mathbf I$ are clustered, each cluster is averaged to create a new $\mathbf I$.

The next step is to group all branches in $\mathbf I$ by their respective topological intersection.  
To do so, we first classify all branches in $\mathbf I$ as either Y or X-branches.
Intersections that are X-branches are already topologically correct so they are left unmodified in $\mathbf I$.
The remaining Y-branches are removed from $\mathbf I$, after which we obtain a list of all possible Y-branch pair combinations and sort them by their pair distance.
The closest branch pairs are then iteratively popped from the list and matched so long as neither branch has already been matched.
A new intersection pixel is then computed from the average of the matched branch locations and then added back to $\mathbf I$.

Using the new intersection pixel location, all matched Y-branches can then be replaced with an X-shaped branch, as shown in Fig.~\ref{fig:intersection_clustering}(c).
Note that there may be cases where an intersection is topologically a Y-branch (i.e., an end perfectly overlaps with a segment) and thus has no corresponding match.
To account for these cases, we stop matching Y-branches once the pair distance exceeds a limit $\epsilon = 10\max(\mathbf D)$ or if every Y-branch has already been matched.
Any remaining non-matched Y-branches are added back to $\mathbf I$.
As shown in Fig.~\ref{fig:intersection_clustering}(c), we record new ``ends'' for all topologically correct intersections. This is done so that we know that an intersection is imminent during a pixel traversal and, hence, take the correct precomputed path, as discussed next. 

\subsection{Minimal Bending Energy Path Generation}

For rods that have nonuniform curvatures, the bending energy must be computed in a discretized fashion.
If we discretize a rod into $N$ nodes and $N-1$ edges, then the total bending energy is 
\begin{equation}
\begin{aligned}
    E_b &= \frac{1}{2} \frac{EI}{V} \sum^{N-2}_{k=1} (\kappa_k - \kappa^0_k)^2,
\end{aligned}
\end{equation}
where $EI$ is the bending stiffness, $\kappa_k$ and $\kappa_k^0$ are the deformed and undeformed discrete dimensionless curvatures, respectively, at node $k \in [1, N-2]$, and $V$ is the Voronoi length.
For our DLOs, we assume that the undeformed curvature is a straight configuration ($\kappa^0 = 0$).
Then, minimizing the bending energy of an elastic rod amounts to minimizing the discrete curvatures.

The norm of the discrete dimensionless curvature for a node $k$ is easily computed using the unit tangent vectors of the adjacent edges \cite{bergou2008der}:
\begin{equation}
    \bar\kappa_k = \left\lVert \frac{2 \mathbf t^{k-1} \times \mathbf t^k}{1 + \mathbf t^{k-1} \cdot \mathbf t^k} \right\rVert,
\end{equation}
where $\mathbf t^{k-1}$ and $\mathbf t^k$ are the unit tangent vectors of edges $k-1$ and $k$, respectively.

Note that the only time we must choose between multiple paths is at an intersection, whereas traversals through segments are unambiguous.
Using the new ends shown in Fig.~\ref{fig:intersection_clustering}(c), we can compute the combination of paths that minimizes the cumulative bending energy of the DLOs by simply computing the pairs of segments that minimize cumulative norm curvature.
In other words, if an intersection at $\mathbf i$ has four end points $\mathbf a, \mathbf b, \mathbf c$, and $\mathbf d$, then we must find the pairs of end points ($\mathbf p_1^1, \mathbf p_1^2$) and ($\mathbf p_2^1, \mathbf p_2^2$) that minimizes $\Vert \kappa_1 \rVert + \lVert \kappa_2 \rVert$, where
\begin{equation}
\begin{split}
    \kappa_1 = \frac{2 \mathbf t_1^1 \times \mathbf t_1^2}{1 + \mathbf t_1^1 \cdot \mathbf t_1^2}, \quad 
    \kappa_2 = \frac{2 \mathbf t_2^1 \times \mathbf t_2^2}{1 + \mathbf t_2^1 \cdot \mathbf t_2^2},\\
    \mathbf t_1^1 = \frac{\mathbf i - \mathbf p_1^1}{\lVert \mathbf i - \mathbf p_1^1 \rVert}, \quad
    \mathbf t_1^2 = \frac{\mathbf p_1^2 - \mathbf i}{\lVert \mathbf p_1^2 - \mathbf i \rVert}, \\
    \mathbf t_2^1 = \frac{\mathbf i - \mathbf p_2^1}{\lVert \mathbf i - \mathbf p_2^1 \rVert}, \quad
    \mathbf t_2^2 = \frac{\mathbf p_2^2 - \mathbf i}{\lVert \mathbf p_2^2 - \mathbf i \rVert}. 
\end{split}
\end{equation}

Fig.~\ref{fig:intersection_clustering}(d) shows an example of this optimization, where out of the 3 possible combinations of paths the one that minimizes total curvature is selected.
With the paths through intersections properly precomputed, the skeleton pixel traversals to obtain each DLO's centerline can now take place.
Algorithm~\ref{alg:mbest_pseudocode} shows the pseudocode of the \textit{mBEST} pipeline.

\begin{algorithm}
\SetAlgoLined
\LinesNumbered
\DontPrintSemicolon
\KwIn{$\mathbf M_\textup{dlo}$}
\KwOut{$\mathbf P$}
\SetKwFunction{Skeletonize}{Skeletonize}
\SetKwFunction{DistTransform}{DistTransform}
\SetKwFunction{ComputeParams}{ComputeParams}
\SetKwFunction{DetectKeyPoints}{DetectKeyPoints}
\SetKwFunction{PruneSplitEnds}{PruneSplitEnds}
\SetKwFunction{ReplaceIntersections}{ReplaceIntersections}
\SetKwFunction{GenIntersectionPaths}{GenIntersectionPaths}
\SetKwProg{Fn}{Func}{:}{}
\SetKwFunction{mBEST}{mBEST}
{
\Fn{\mBEST{$\mathbf M_\textup{dlo}$}}
{
$\mathbf P \gets [ \ ]$\;
$\mathbf M_\textup{sk} \gets \Skeletonize(\mathbf M_\textup{dlo})$\;
$\mathbf D \gets \DistTransform(\mathbf M_\textup{dlo})$\;
$\mathbf \delta, \epsilon \gets \ComputeParams(\mathbf D)$\;
$\mathbf E, \mathbf I \gets \DetectKeyPoints(\mathbf M_\textup{sk} \circledast \mathbf K$)\;
$\mathbf E, \mathbf I \gets \PruneSplitEnds(\mathbf E, \mathbf I, \mathbf M_\textup{sk}, \delta)$\;
$\mathbf I \gets \ReplaceIntersections(\mathbf I, \epsilon)$\;
$\mathbf P_\textup{inter} \gets \GenIntersectionPaths(\mathbf I, \mathbf M_\textup{sk})$\;
\While{$\mathbf E$ \textup{is not empty}} {
    $\mathbf x \gets \mathbf E$.pop()\;
    \While{\textup{True}} {
        $\boldsymbol \tau \gets$ traverse along $\mathbf M_\textup{sk}$ from $\mathbf x$ until reaching an end $\mathbf e$\;
        \If{$\mathbf e \in \mathbf P_\textup{inter}$} {
            $\boldsymbol \tau \gets \boldsymbol \tau + \mathbf P_\textup{inter}^i$\;
            $\mathbf x \gets $ last pixel of $\mathbf P_\textup{inter}^i$\;
        }
        \Else{
            $\mathbf E$.remove$(\mathbf e)$\;
            break\;
        }
    }
    $\mathbf P$.append($\boldsymbol \tau$)\;
}
\textbf{return} $\mathbf P$ \;
}
}
\caption{\textit{mBEST} Pipeline Pseudocode}
\label{alg:mbest_pseudocode}
\end{algorithm}

\subsection{Crossing Order Determination}

The final step of the pipeline involves ascertaining which DLO is resting on top at intersections. 
To solve this problem, we use a modified version of \textit{FASTDLO}'s  \cite{caporali2022fastdlo} solution.
To compute crossing order at intersections, we use the precomputed optimal paths shown in Fig.~\ref{fig:intersection_clustering}(d). 
Crossing order is then determined by computing the sum of the standard deviations of the RGB channels of the pixels along each path.
Finally, the path that contains the lower sum is assumed to be the one on top.
Although this solution from \textit{FASTDLO} works fairly well, we discovered that failures can occur due to glare along the centerline, which may even cause failures for intersections with two completely different colored DLOs.
To eliminate the influence of glare, we compute the standard deviations of the intersection path pixels not on the original input image but on its blurred version.

\section{Experimental Results}
\label{sec:exp}

\begin{figure*}
	\includegraphics[width=\textwidth]{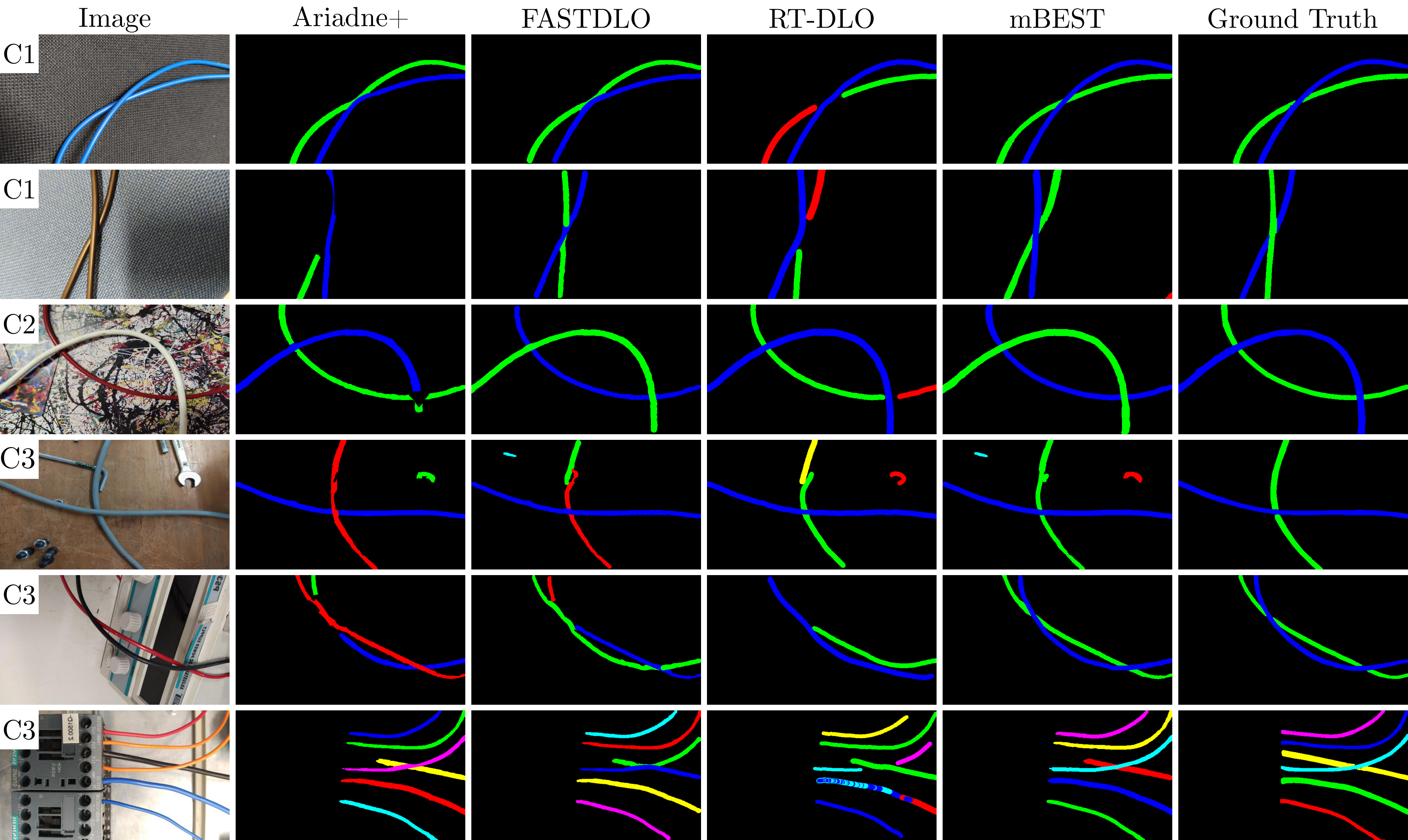}
	\caption{Sample segmentations for the simple configuration against complex background dataset. Each row shows segmentation results for a different image with the left column indicating the dataset to which the image belongs. Columns~2--5 show \textit{Ariadne+}, \textit{FASTDLO}, \textit{RT-DLO}, and \textit{mBEST} results, respectively. The right column shows the ground truth. Note the failure to properly handle intersections for all baseline algorithms, especially when strands are nearly parallel. In fact, \textit{RT-DLO} can be seen to produce an unintuitive output for the last example where certain wires are labeled multiple times.}
	\label{fig:complex_background_results}
\end{figure*}

\begin{figure*}
	\includegraphics[width=\textwidth]{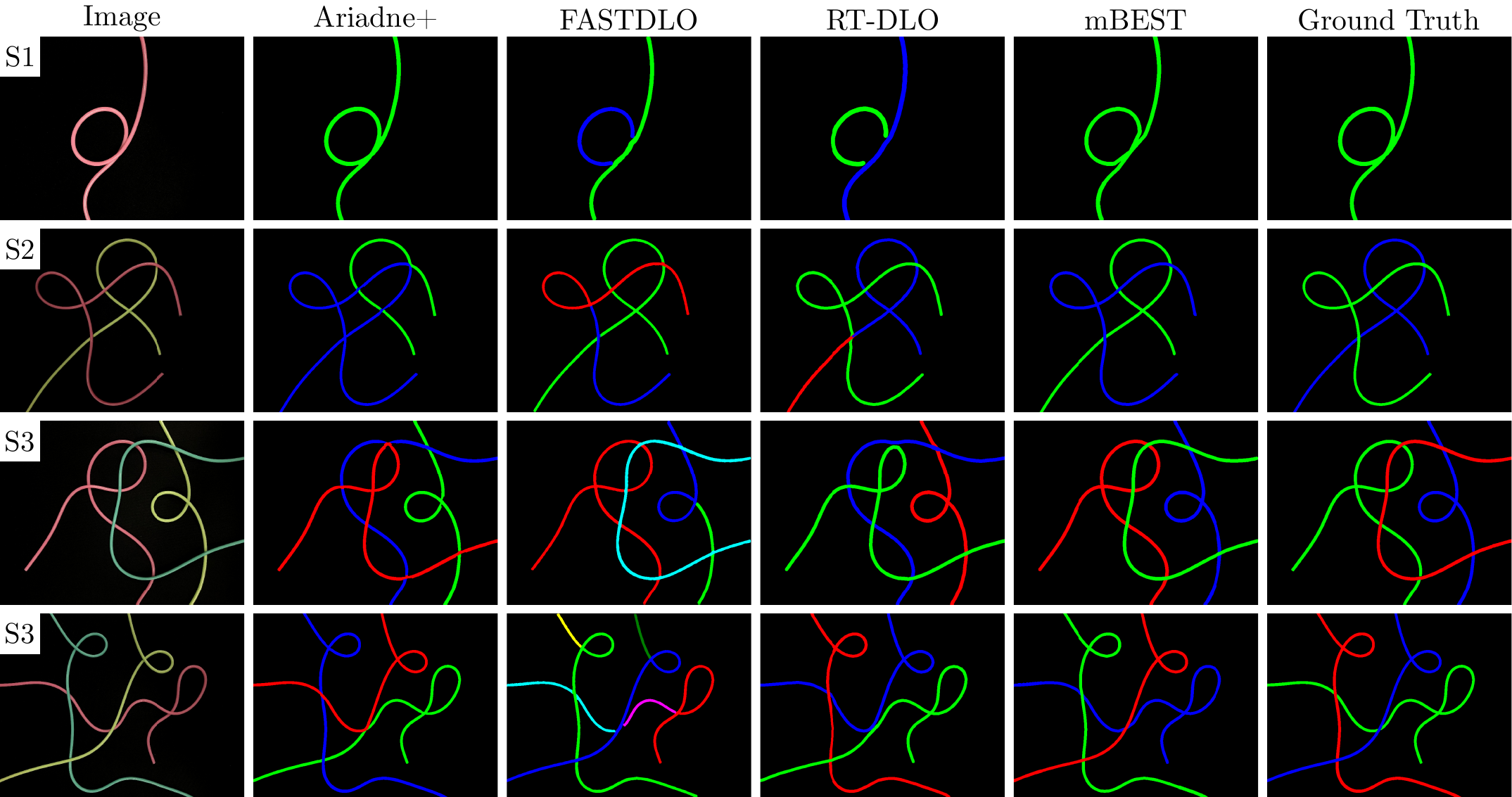}
	\caption{Sample segmentations for the complex configuration against simple background dataset. Each row shows segmentation results for a different image with the left column indicating the dataset to which the image belongs. Columns~2--5 show results for \textit{Ariadne+}, \textit{FASTDLO}, \textit{RT-DLO}, and \textit{mBEST}, respectively. The right column shows the ground truth. Several cases of incorrect intersection handling can be observed for all the baseline algorithms, whereas \textit{mBEST} robustly handles intersections using its simple bending energy optimization.}
	\label{fig:simple_background_results}
\end{figure*}

\subsection{Datasets}

We used two different datasets to evaluate the effectiveness of \textit{mBEST}. The first consists of relatively simple configurations of DLOs against complex backgrounds, whereas the second consists of complex configurations (i.e., highly varying curvatures and numerous self-loops) of DLOs against a simple black background. We focused mostly on images with a simple black background since the initial binary mask segmentation is not a key aspect of our algorithm; however, \textit{mBEST} also works well for complex backgrounds, as shown in Fig.~\ref{fig:complex_background_results}.

The complex background dataset was provided by \cite{caporali2023rtdlo}, and comprises a total of 132 images of size $640\times360$. 
It is split into tiers C1, C2, and C3, each containing 44 images, where the tier numbers reflect the increasing complexity of the background.
Given the complexity of the background, DCNN segmentation was used to obtain the initial binary mask.
We removed two images each from C2 and C3 as they included intersections involving $> 2$ DLOs, scenarios which are currently outside the scope of \textit{mBEST}.

The simple background dataset consists of a total of 300 images of size $896 \times 672$ and is split into tiers S1, S2, and S3, each containing 100 images, where the tier numbers reflect the number of DLOs in the image, resulting in both an increase in complexity and computational demand as the numbers increase.
Given the high contrast background, color filtering sufficed to obtain the initial binary mask.

\subsection{Baselines and Parameters}

We tested \textit{mBEST} against three state-of-the-art baselines: \textit{Ariadne+}~\cite{caporali2022ariadne+}, \textit{FASTDLO}~\cite{caporali2022fastdlo}, and \textit{RT-DLO}~\cite{caporali2023rtdlo}.
In terms of hyperparameters, the number of superpixels for \textit{Ariadne+} was set to 75 for complex background images and to 200 for simple background images.
Both these values were chosen as optimal after performing a parameter sweep on each dataset. 
For \textit{RT-DLO}, the K-nearest neighbors matching parameter was set to 8, the edge similarity threshold was set to 0.1, and the vertex sampling ratio was set to 0.15.
These hyperparameters were provided by default and shown to have good performance in \cite{caporali2023rtdlo}. 
For all experiments involving use of the DCNN model, a pixel segmentation threshold of 77 (0--255) was used.
Furthermore, although \textit{Ariadne+} has its own neural network for the initial segmentation of the DLOs, we replaced it with \textit{FASTDLO}'s DCNN model for consistency and better performance.

Additionally, we demonstrated the effectiveness of \textit{mBEST}'s skeleton refinement steps by conducting experiments on an aggregated dataset consisting of S1, S2, and S3 with a hybrid formulation that uses \textit{FASTDLO}'s intersection handling neural network in \textit{mBEST}'s framework.

All experiments were run on a workstation with an Intel i9-9900KF CPU and an NVIDIA RTX 2080 Ti GPU.

\subsection{Results and Analysis}

\begin{table*}
\setlength{\tabcolsep}{8.75pt}
\renewcommand{\arraystretch}{1.2}
\centering
\caption{Experimental Results}
\begin{tabular}{c cccc cccc}
\toprule
\multirow{2}{*}[-3pt]{Dataset } &  \multicolumn{4}{c}{DICE [\%]} & \multicolumn{4}{c}{Runtime [FPS]}  \\
\cmidrule(lr){2-5}\cmidrule(lr){6-9}
& \textit{Ariadne+} & \textit{FASTDLO} & \textit{RT-DLO} & \textit{mBEST} & \textit{Ariadne+} & \textit{FASTDLO} & \textit{RT-DLO} & \textit{mBEST} \\
\midrule
C1 & $88.30\pm0.102$ & $89.82\pm0.091$ & $90.31\pm0.085$ & $\mathbf{91.08}\pm\mathbf{0.083}$ & $2.69$ & $20.81$ & $30.58$ & $\mathbf{31.86}$ \\
C2  & $91.03\pm0.044$ & $91.45\pm0.039$ & $91.10\pm0.058$ & $\mathbf{92.17\pm0.050}$ & $2.63$ & $20.90$ & $\mathbf{32.50}$ & $32.03$ \\
C3  & $86.13\pm0.123$ & $86.55\pm0.110$ & $87.27\pm0.128$ & $\mathbf{89.69\pm0.089}$ & $2.72$ & $20.51$ & $\mathbf{32.44}$ & $32.17$ \\
S1  & $97.24\pm0.065$ & $87.91\pm0.062$ & $96.72\pm0.014$ & $\mathbf{98.21}\pm\mathbf{0.006}$ & $0.92$ & $21.88$ & $39.60$ & $\mathbf{52.79}$ \\
S2  & $96.81\pm0.074$ & $88.92\pm0.061$ & $94.91\pm0.019$ & $\mathbf{97.10\pm0.010}$ & $0.78$ & $17.34$ & $25.73$ & $\mathbf{41.04}$ \\
S3  & $96.28\pm0.067$ & $90.24\pm0.042$ & $94.12\pm0.043$ & $\mathbf{96.98\pm0.009}$ & $0.73$ & $15.33$ & $22.06$ & $\mathbf{37.11}$ \\
\bottomrule
\end{tabular}
\label{tab:dice_results}
\end{table*}

We report two key metrics.
First, we look at segmentation accuracy using the popular DICE metric.
We also report the average run times for each algorithm in frames per second (FPS).
Table~\ref{tab:dice_results} reports both metrics for all our experiments.

For the complex background datasets, we see that \textit{mBEST} outperforms all baseline algorithms in terms of mean DICE score.
In particular, we see that the baseline algorithms often struggle to handle intersections that are nearly parallel, as shown in Fig.~\ref{fig:complex_background_results}.

With regard to runtime, \textit{mBEST} is roughly on par with \textit{RT-DLO} and is a clear improvement over \textit{Ariadne+} ($\approx 11\times$) and \textit{FASTDLO} ($\approx 1.5\times$).
Note three important caveats for these results: 1) the initial DCNN segmentation comprises a large portion of the computation time; 2) the images are relatively small and the number $N$ of DLOs is random, giving little insight as to how the algorithms scale with $N$, and 3) \textit{RT-DLO}'s ability to keep up with \textit{mBEST} in speed is solely due its low vertex sampling rate (0.15).
We observe that increasing the sampling rate increases the compute time significantly given the computational expense of graph construction.

To address the above concerns, consider the results for the simple background datasets.
As these datasets do not require the use of a DCNN and are labeled by the number of DLOs they contain, we can accurately determine how each algorithm scales and performs with respect to $N$.
As reported in the bottom half of Table~\ref{tab:dice_results}, \textit{mBEST} offers clear speed improvements over the \textit{Ariadne+} ($\approx 54\times$), \textit{FASTDLO} ($\approx 2.4\times$), and \textit{RT-DLO} ($\approx 1.5\times$) baselines.
Additionally, we see that \textit{mBEST} scales better with respect to $N$ compared to \textit{RT-DLO} despite the latter's sparse sampling rate, with \textit{mBEST} experiencing runtime decreases of about $22.3\%$ and $9.6\%$ when moving up each tier compared to \textit{RT-DLO}'s $35\%$ and $14.3\%$.
Though a low sampling rate works well for the relatively straight configurations of DLOs in C1, C2, and C3, we notice that performance degrades significantly once a coarse sampling rate is unable to capture the highly variable curvatures of complex assemblies of rods (i.e., those in S1, S2, and S3).
Examples of this can be observed in our supplementary video (see Footnote~\ref{github}).

In addition to the significant improvement in runtime, \textit{mBEST} also outperforms all the baseline algorithms in terms of mean DICE score as well. 
Several examples of intersection failures experienced by the baseline algorithms are shown in Fig.~\ref{fig:simple_background_results}.
Such failures typically occur in extreme cases (i.e., either nearly-parallel or extremely curved self-loops).
Interestingly, \textit{Ariadne+}'s mean DICE score is very close to \textit{mBEST}'s, but had up to $10\times$ the standard deviation, meaning that \textit{Ariadne+} suffered a higher number of outright failures. 
In fact, \textit{mBEST} has a lower standard deviation compared to all the baseline algorithms across all the datasets with the exception of C2, indicating a higher level of consistency for a wide range of data.

\begin{table}
\renewcommand{\arraystretch}{1.2}
\setlength{\tabcolsep}{10.5pt}
\centering
\caption{Skeleton Refinement Analysis on S1+S2+S3}
\begin{tabular}{lcc}
\toprule
\textbf{Algorithm} & DICE [\%] &  Runtime [FPS] \\
\midrule
\textit{FASTDLO} & $89.02 \pm 0.056$ & $16.13$ \\
\textit{HYmBEST}  & $97.39 \pm 0.013$ & $29.94$ \\
\textit{mBEST}  & $97.43 \pm 0.010$& $42.86$ \\
\bottomrule
\end{tabular}
\label{tab:skeleton_refinement_analysis}
\end{table}

Finally, we analyze the effectiveness of our skeleton refinement steps by formulating a hybrid algorithm, \textit{HYmBEST}, which uses \textit{FASTDLO}'s intersection handling neural network (IHNN) plugged into \textit{mBEST}'s framework.
As reported in Table~\ref{tab:skeleton_refinement_analysis}, \textit{HYmBEST} achieves a mean DICE score almost identical to \textit{mBEST}, with both significantly outperforming \textit{FASTDLO}.
This is noteworthy as it shows that \textit{FASTDLO}'s IHNN works reasonably well, but that the improperly handled skeleton structure yields poor results, thus highlighting the importance of the topology-correcting refinement steps.
Note also that although \textit{FASTDLO}'s IHNN can perform well in a hybrid formulation, \textit{mBEST}'s remarkably cheap bending energy formulation still results in an $\approx 43\%$ runtime improvement.

\section{Conclusion}
\label{sec:conclusion}

We have introduced \textit{mBEST}, an end-to-end pipeline for the segmentation of deformable linear objects (DLOs) in images that improves upon the state of the art both in terms of accuracy and computational speed.
Through a variety of experiments, we have shown that \textit{mBEST} can robustly handle complex scenes with highly tangled DLOs by generating paths on topologically correct skeletons that minimize the cumulative bending energy of the scene.

In future work, we will explore solutions that take into consideration occlusions, multiple DLOs at an intersection, poor quality binary masks, and dense knots; i.e., strands touching in parallel.
We note that though we do not cover it in this manuscript, the bending energy formulation of \textit{mBEST} can easily be expanded to deal with multiple DLOs at an intersection by simply accounting for additional path combinations. 
Furthermore, methods like \textit{RT-DLO}~\cite{caporali2023rtdlo} already take into consideration the possibility of poor binary masks and may be better suited for such situations. 
Finally, another promising research direction is 3D detection of DLOs, thus enabling robots to go beyond simple planar manipulation.
Solutions for this may involve using \textit{mBEST} to generate segmentations from multiple viewing angles for the purposes of 3D reconstruction.


\balance 
\bibliographystyle{ieeetr}
\bibliography{main_arxiv_v3}

\end{document}